\documentclass[11pt]{article}

\usepackage{arxiv}
\usepackage[utf8]{inputenc}
\usepackage[T1]{fontenc}
\usepackage{amsmath}
\usepackage{amssymb}
\usepackage{booktabs}
\usepackage{multirow}
\usepackage{graphicx}
\usepackage{subcaption}
\usepackage{float}
\usepackage{placeins}
\usepackage{xspace}
\usepackage{microtype}
\usepackage[numbers,sort&compress]{natbib}
\usepackage{hyperref}

\newcommand{\method}{PACE\xspace}

\providecommand{\Description}[1]{}

\title{PACE: Polar Axis-Conditioned Estimation for PairUAV Relative Localization}

\author{
  Ze Rong\\
  FST and ICI, University of Macau\\
  Macau, China
}

\date{}
\renewcommand{\headeright}{Preprint}
\renewcommand{\undertitle}{Preprint}
\renewcommand{\shorttitle}{PACE}

\hypersetup{
  colorlinks=true,
  linkcolor=blue,
  citecolor=blue,
  urlcolor=blue,
  pdftitle={PACE: Polar Axis-Conditioned Estimation for PairUAV Relative Localization},
  pdfauthor={Ze Rong},
  pdfkeywords={UAV localization, relative pose estimation, PairUAV, axis-conditioned estimation}
}

\begin{document}
\maketitle

\begin{abstract}
PairUAV relative localization maps two UAV images to a polar navigation
command. Although heading and range share the same pairwise pose context,
treating them as homogeneous coordinates forces both outputs to use the same
decoder evidence and optimization state. Controlled readout probes reveal a
different structure: the two axes favor different decoder-depth combinations,
their best checkpoints disagree on 80.8\% of a validation trajectory, and
range errors exhibit a distinct high-error tail. We introduce \method, Polar
Axis-Conditioned Estimation, which retains a shared Reloc3r-style pair
representation while assigning axis-specific readout interfaces. Heading uses
mid/late relational evidence, whereas range remains attached to a direct late
metric path. On the official hidden test, the strongest released raw predictor
scores 0.002460; the complementary PAAER predictor scores 0.002514 with a
slightly lower angle error. Deterministic challenge packaging, reported
separately from learned estimation, yields the final score of 0.001874. Code,
checkpoints, predictions, and reconstruction tools are available at
\url{https://github.com/zerong7777-boop/PairUAV-PACE}.
\end{abstract}


\section{Introduction}
\label{sec:introduction}

PairUAV relative localization, introduced in the UAVM challenge
line~\cite{deuser2026UVA,li2026lastmeterprecisionnavigationuavs,pairuav2026challenge},
asks a model to infer a polar navigation command from two UAV-view images: a
heading $\theta$ and a metric range $d$. Both quantities describe the same
pairwise pose, making end-to-end joint regression a natural formulation. Yet
the usual map $(I^a,I^b)\mapsto(\theta,d)$ hides a consequential design choice:
sharing pair context does not require the two outputs to consume the same
decoder evidence or to rely on the same point along training.

Several localization routes can be adapted to this task. Cross-view retrieval
ranks candidate places rather than directly predicting a command
~\cite{zhu2022transgeotransformerneedcrossview}; dense matching and feed-forward
geometry expose correspondence and 3D structure
~\cite{sun2021loftrdetectorfreelocalfeature,edstedt2023romarobustdensefeature,wang2024dust3rgeometric3dvision,leroy2024groundingimagematching3d,wang2025vggtvisualgeometrygrounded},
but require an additional mapping to the two scored scalars. Reloc3r-style
relative pose regression (RPR) instead maps an image pair directly to a
low-dimensional relative pose~\cite{dong2025reloc3rlargescaletrainingrelative}.
Its output contract and strong adaptation results make it a suitable shared
host for PairUAV. The remaining question is how the host should expose its
decoder evidence to the two polar axes.

Our development probes reveal \emph{axis-conditioned reliability}. Under a
matched protocol, heading and range respond differently when early, middle,
and late decoder features are exposed to the readout. Along a validation
checkpoint trajectory, the heading-best and range-best states disagree for
80.8\% of image pairs; a separate 4,096-pair surface reproduces the trend at
77.3\%. A tested geometry-assisted source likewise favors heading on many
pairs while weakening range, and a competitive range median can coexist with a
severe high-error tail. These observations do not imply independent latent
factors. Representation controls instead show that both pose regimes remain
readable from shared features. The failure lies in treating evidence utility
and optimization reliability as homogeneous after that shared context is
formed.

We therefore introduce \method, Polar Axis-Conditioned Estimation. \method
retains a shared Reloc3r-style encoder--decoder and changes the readout
contract: heading may aggregate mid/late relational evidence, whereas range
remains attached to a direct late metric interface. Multi-Depth Heading
Readout (MDHR) and Protected Axis-Asymmetric Expert Readout (PAAER) instantiate
this principle with different heading interfaces. The learned estimators are
evaluated independently. Range fusion and structured output calibration are
reported separately as deterministic challenge packaging.

\textbf{Contributions.} (1) We identify axis-conditioned reliability in
PairUAV through matched readout controls and cross-surface trajectory
diagnostics. (2) We propose \method, which preserves shared pair context while
allocating decoder evidence by polar axis. (3) We evaluate the learned
estimators and the deterministic submission protocol separately, and release
the artifacts required to reconstruct the reported challenge result.

\section{Related Work}
\label{sec:related}

\subsection{Pairwise Localization and Relative Pose Regression}

Cross-view geo-localization commonly retrieves a place across aerial and
ground viewpoints~\cite{zheng2020university1652,zhu2022transgeotransformerneedcrossview},
with recent aerial extensions incorporating multi-scale and textual cues
~\cite{wang2024museaerial,chu2024geotext1652}. PairUAV differs in its output:
each image pair must yield a continuous navigation command rather than a ranked
location. Relative pose regression is therefore a closer reference. Reloc3r
uses a symmetric shared-weight image-pair network to regress relative camera
pose before optional motion averaging~\cite{dong2025reloc3rlargescaletrainingrelative}.
We adopt this efficient pairwise host, but revisit its terminal contract because
PairUAV evaluates heading and metric range as separate polar axes.

\subsection{Matching and Feed-Forward Geometry}

LoFTR and RoMa estimate dense correspondences directly from image content
~\cite{sun2021loftrdetectorfreelocalfeature,edstedt2023romarobustdensefeature}.
DUSt3R regresses dense pointmaps, MASt3R augments them with descriptors for
3D-grounded matching, and VGGT predicts cameras, depth, points, and tracks in a
feed-forward model~\cite{wang2024dust3rgeometric3dvision,leroy2024groundingimagematching3d,wang2025vggtvisualgeometrygrounded}.
These outputs are well suited to correspondence and coordinate-frame geometry,
but they are not themselves the two PairUAV scalars. We use them as candidate
evidence sources and adaptation controls. Their heterogeneous axis profiles
motivate selective evidence use rather than indiscriminate feature
concatenation.

\subsection{Shared Representations with Task-Specific Readouts}

Multi-task learning balances shared representations against negative transfer.
Uncertainty weighting, gradient normalization, and gradient surgery adjust how
objectives interact during optimization
~\cite{kendall2018multitasklearningusinguncertainty,chen2018gradnormgradientnormalizationadaptive,yu2020gradientsurgerymultitasklearning}.
PairUAV presents a narrower case: heading and range are coupled coordinates of
one relative command, not unrelated tasks. Full separation discards useful
pair context, whereas a homogeneous output path can expose both axes to
evidence that benefits only one. \method addresses this middle ground through
a shared pair representation and axis-conditioned decoder readouts.

\section{Method}
\label{sec:method}

\subsection{Task and Homogeneous-Readout Assumption}
\label{sec:method_formulation}

Given $x_i=(I_i^a,I_i^b)$, PairUAV predicts
$y_i=(d_i,\theta_i)$, where $d_i$ is range and $\theta_i$ is heading. The
submission file uses \texttt{angle distance} order, although the equations use
range--heading order. The official evaluator reports range relative error,
angle relative error, and their mean. For nonzero targets,
\begin{equation}
 e_i^d=\frac{|\hat d_i-d_i|}{|d_i|},\qquad
 e_i^\theta=\frac{\Delta(\hat\theta_i,\theta_i)}{|\theta_i|},
 \label{eq:official_metric}
\end{equation}
where $\Delta$ is the circular absolute difference after normalization. For a
zero denominator, an exact zero prediction receives zero error; otherwise that
axis term is excluded according to the official rule. The final score averages
$\frac{1}{2}(e_i^d+e_i^\theta)$ over valid terms.

Let a Reloc3r-style encoder--decoder produce a feature bank
$\mathcal Z=\{z^{(1)},\ldots,z^{(L)}\}$. A homogeneous readout predicts both
axes from one terminal interface,
$(\hat d,\hat v_\theta)=H(z^{(L)})$. This shares context and evidence selection
simultaneously. The latter is unnecessary: decoder features that improve
heading need not preserve range reliability.

\subsection{Polar Axis-Conditioned Readout}
\label{sec:axis_readout}

\method separates shared representation from axis-specific evidence
allocation. For axis $a\in\{d,\theta\}$, it selects a decoder subset
$\mathcal S_a$ and applies
\begin{equation}
 u_a=G_a(\{z^{(\ell)}:\ell\in\mathcal S_a\}),\qquad
 \hat y_a=R_a(u_a).
 \label{eq:axis_conditioned}
\end{equation}
The range interface uses $\mathcal S_d=\{L\}$, while heading may use middle
and late layers. The backbone remains shared; \method does not assert latent
factor independence.

Multi-Depth Heading Readout (MDHR) retains the terminal range path and
aggregates selected intermediate and final decoder features only for heading.
The matched depth controls in Sec.~\ref{sec:readout_evidence} determine which
layers are exposed. This design tests evidence allocation rather than simply
increasing both heads.

Protected Axis-Asymmetric Expert Readout (PAAER) is a complementary
instantiation. It predicts range directly as
$\hat d^P=R_d^P(z^{(L)})$ and forms the heading state with query-bridge
attention,
$u_\theta^P=A_\theta(q_\theta,\{z^{(m)},z^{(L)}\})$. The heading head outputs
a normalized vector $\hat v_\theta$ before conversion with \texttt{atan2}.
Here ``expert'' means an axis-specialized readout, not mixture-of-experts
routing. Range is preserved at the readout interface only: because the shared
backbone is jointly fine-tuned, heading gradients can still alter late
features.

\subsection{Optimization}
\label{sec:training_objective}

Let $\delta_i^\theta=\Delta(\hat\theta_i,\theta_i)$ and
$\delta_i^d=|\hat d_i-d_i|$. The full-scale runs use the metric-aware objective
\begin{align}
 L_\theta &= \mathbb E\!\left[\frac{\delta_i^\theta}
 {\max(q_\theta(\theta_i),1^\circ)}
 +0.05\,\operatorname{SmoothL1}(\delta_i^\theta,0)\right],\\
 L_d &= \mathbb E\!\left[\frac{\delta_i^d}{\max(|d_i|,1)}
 +\gamma_d\,\operatorname{SmoothL1}(\delta_i^d,0)\right],\qquad
 L=L_\theta+L_d,
 \label{eq:training_loss}
\end{align}
where $q_\theta$ follows the implementation's normalized heading magnitude.
MDHR uses $\gamma_d=0.05$. The final PAAER continuation uses $\gamma_d=0.10$
and multiplies both range terms by
$w_i=1+2\,\operatorname{clip}((|d_i|-80)/40,0,1)$, reaching weight 3 at
$|d_i|\ge120$. No validation labels define these sample weights.

Models start from Reloc3r-512 checkpoints and use $512{\times}384$ inputs,
AdamW, mixed precision, batch size 4, learning rate $10^{-5}$, no warmup, and
full-model fine-tuning. The scaled MDHR and final PAAER runs each continue for
one full train epoch; the latter starts from the 10k tail-aware checkpoint.
Released scripts fix the data manifests, initialization, and seeds.

\subsection{Deterministic Challenge Packaging}
\label{sec:challenge_packaging}

The final submission keeps PAAER heading and fuses three released range
predictions:
\begin{equation}
 \hat d^{F}=0.511\hat d^{P}+0.189\hat d^{M}+0.300\hat d^{B},
 \label{eq:rsf}
\end{equation}
where $M$ and $B$ denote MDHR and the metric-aware Reloc3r baseline. The
coefficients were selected during challenge-time submission sweeps using
aggregate leaderboard feedback; individual hidden-test labels were never
available. This fusion is therefore packaging, not a learned module or a
general fusion law.

Finally, heading is rounded to the nearest $2^\circ$ lattice and range is
projected to the nearest value in the public train/dev support
$\mathcal S_d$:
\begin{equation}
 \tilde\theta=2\operatorname{round}(\hat\theta^P/2),\qquad
 \tilde d=\arg\min_{s\in\mathcal S_d}|s-\hat d^F|.
 \label{eq:structured_projection}
\end{equation}
These deterministic operations exploit PairUAV output structure and do not
participate in training. They must be revalidated when the target output space
changes.

\section{Experiments}
\label{sec:experiments}

\subsection{Setup and Evidence Surfaces}

We use three evidence levels. Official results are aggregate hidden-test
feedback from the challenge server. Architecture probes use fixed local
surfaces and are compared only within matched protocol blocks. Mechanism
diagnostics use labeled validation or train-hash trajectories and are not
inference-time selectors. We report circular heading MAE in degrees, absolute
range MAE in dataset distance units, and a lower-is-better local proxy aligned
with the two official relative errors.

\subsection{Host Choice and Readout Evidence}
\label{sec:readout_evidence}

Bounded adaptation probes first compared candidate evidence sources. RoMa
obtained heading/range MAE of 15.39/19.47, MASt3R geometry 9.59/33.23, and VGGT
geometry 52.51/14.17. These mixed-protocol diagnostics are not a global model
ranking, but their different axis profiles and weaker command regression
motivated a direct RPR host. Reloc3r-family models subsequently produced all
strong official raw predictors in Table~\ref{tab:official_results}.

Table~\ref{tab:readout_depth} tests the central design choice under matched
Phase88 blocks. At 2.5k steps, exposing every tested depth lowers heading MAE
but degrades range enough to trail mid--late in the joint proxy. Mid--late also
outperforms the full-depth readout after the matched one-epoch continuation.
Thus, the gain is evidence selection rather than monotonic benefit from more
decoder features.

\begin{table*}[t]
  \centering
  \caption{Matched readout-depth controls. Each budget block uses the same
  data surface, initialization, optimizer, and loss. Heading MAE is in degrees;
  range MAE is in dataset distance units. Lower is better.}
  \label{tab:readout_depth}
  \footnotesize
  \begin{tabular*}{\textwidth}{@{\extracolsep{\fill}}llrrrr@{}}
    \toprule
    Readout & Decoder evidence & Budget & Heading MAE & Range MAE & Proxy \\
    \midrule
    Late only & late & 2.5k & 2.3206 & 4.8927 & 0.015713 \\
    Mid only & middle & 2.5k & 1.9968 & \textbf{4.2571} & 0.013609 \\
    Early--late & early + late & 2.5k & \textbf{1.7596} & 4.8632 & 0.014098 \\
    MDHR & middle + late & 2.5k & 1.9028 & 4.2976 & \textbf{0.013425} \\
    Full-depth & early + middle + late & 2.5k & 1.7947 & 5.1675 & 0.014772 \\
    \midrule
    MDHR & middle + late & 1 epoch & \textbf{1.2113} & \textbf{3.4327} & \textbf{0.009866} \\
    Full-depth & early + middle + late & 1 epoch & 1.2808 & 3.6639 & 0.010497 \\
    \bottomrule
  \end{tabular*}
  \vspace{1pt}

  \parbox{\textwidth}{\scriptsize\textit{Cost audit.} Host: 449.4M parameters;
  official inference: 7.5--8.1 GB at batch 16; MDHR training:
  0.26--0.27 s/step and 25.5 GB peak at batch 4.}
\end{table*}

A separate capacity control is also negative: enlarging the heading branch
changes heading/range MAE from 1.7925/4.5583 to 1.8504/4.6857. For PAAER, the
balanced full-loss readout reaches 1.4985/3.9690 and proxy 0.011680, compared
with 2.3063/3.5898 and 0.013205 without the late-path-preserving interface.
The result supports axis balance, not an improvement in average range from
interface preservation alone; range-tail behavior is analyzed below.

\subsection{Axis-Conditioned Reliability}

Representation controls clarify the sharing boundary. On val811, the
true-minus-shuffled within-regime similarity gaps are 0.609 for heading, 0.258
for absolute range, and 0.590 for signed range. Both axes are therefore
readable from a shared representation; the evidence does not support hard
latent disentanglement.

\begin{figure}[t]
  \centering
  \includegraphics[width=\linewidth]{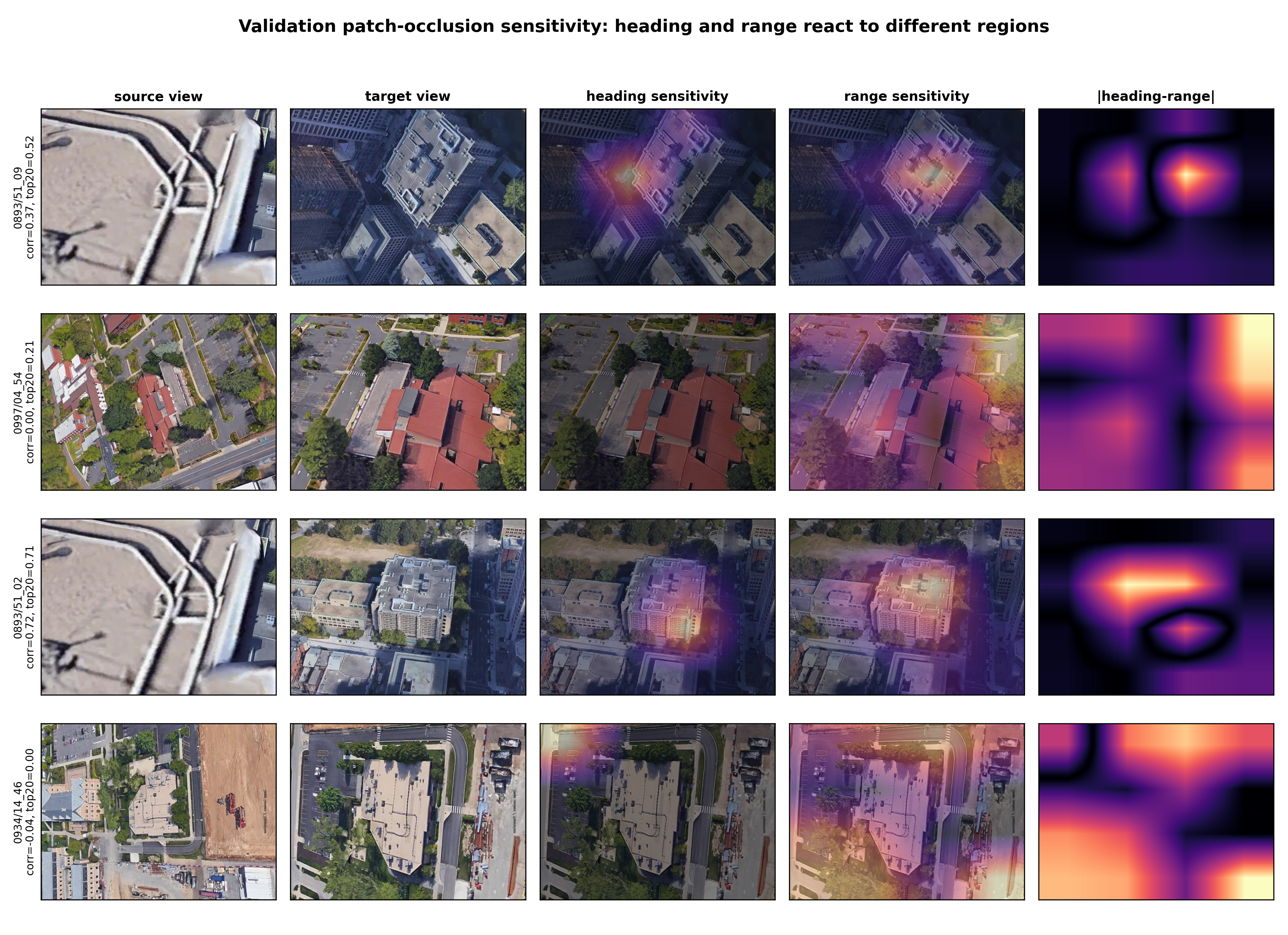}
  \caption{Patch-occlusion sensitivity on four selected validation pairs.
  Heading and range can respond to different target-view regions within the
  same MDHR model. The cases are qualitative diagnostics, not a dataset-level
  causal test.}
  \Description{Four validation image pairs with source and target views,
  heading and range patch-occlusion sensitivity maps, and their difference.}
  \label{fig:axis_evidence}
\end{figure}

A source-level diagnostic provides complementary evidence. On the same 811
validation pairs, one geometry-assisted source improves heading on 694 pairs
but worsens range on 758; 645 pairs lie in the heading-helped/range-harmed
quadrant. This result concerns that source and protocol, not geometry in
general. Figure~\ref{fig:axis_evidence} instead probes one MDHR model by
occluding target-view patches and measuring axis-specific output changes.

Checkpoint trajectories expose an optimization counterpart. On val811,
heading-best and range-best checkpoints differ for 655 of 811 pairs (80.8\%). A
separate 4,096-pair training surface gives 77.3\%, while range-best states occur
on average 31.6k and 36.5k steps earlier, respectively. These
label-assisted winners quantify training asynchrony; they are not a deployable
checkpoint-selection rule.

\begin{figure}[t]
  \centering
  \includegraphics[width=0.84\linewidth]{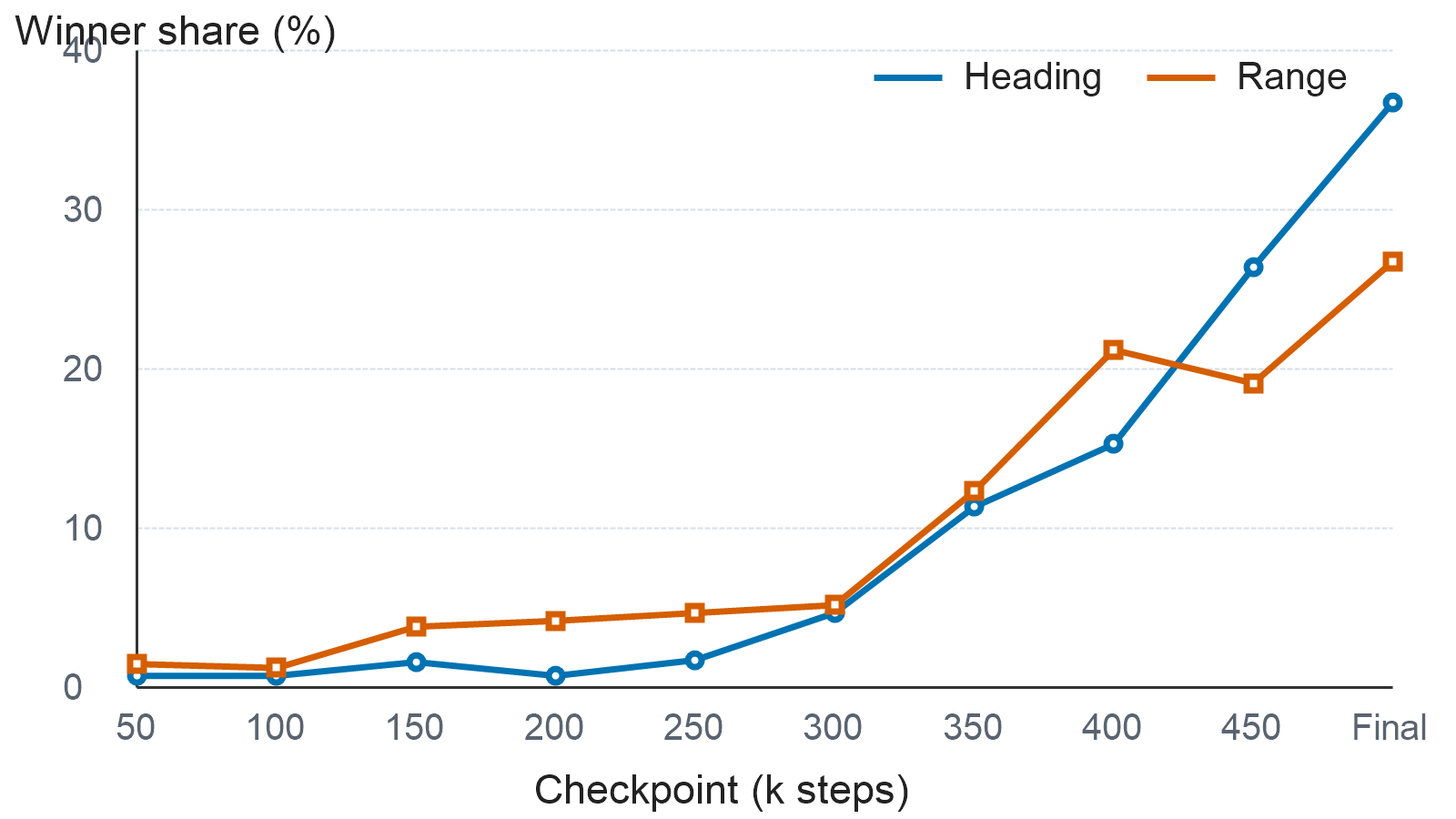}
  \caption{Per-axis best-checkpoint shares on val811. Heading and range select
  different checkpoints for 80.8\% of pairs; an independent training surface
  reproduces the mismatch at 77.3\%.}
  \Description{Line plot of validation samples for which each saved checkpoint
  minimizes heading or range error.}
  \label{fig:checkpoint_axis}
\end{figure}

Range tails form a third reliability axis. Table~\ref{tab:range_tail} compares
a 50k PAAER-line checkpoint, an MDHR checkpoint at 50k, and a further 10k
tail-weighted continuation. The continuation is not a same-budget architecture
comparison; it tests whether the diagnosed failure is repairable. It reduces
the PAAER-line p95, p99, and maximum errors by 85.3\%, 97.1\%, and 95.2\%.

\begin{table}[!t]
  \centering
  \caption{Absolute range-error quantiles on the local diagnostic surface.}
  \label{tab:range_tail}
  \footnotesize
  \begin{tabular}{lrrrr}
    \toprule
    Run & Median & P95 & P99 & Max \\
    \midrule
    PAAER 50k & 0.4546 & 7.1223 & 42.9710 & 91.2856 \\
    MDHR 50k & 0.3897 & 1.1979 & 2.1321 & 6.9968 \\
    PAAER + tail 10k & \textbf{0.3600} & \textbf{1.0446} & \textbf{1.2653} & \textbf{4.3494} \\
    \bottomrule
  \end{tabular}
\end{table}

\subsection{Official Challenge Results}

Table~\ref{tab:official_results} separates raw learned predictors from the
deterministic final package. MDHR is the strongest released raw model by final
score. PAAER is not a stronger overall learner, but has a slightly lower angle
relative error than MDHR (0.002636 vs. 0.002646) and supplies the final heading
source. The final score of 0.001874 belongs to the fused and calibrated package,
not to either readout alone.

\begin{table}[!t]
  \centering
  \caption{Official hidden-test results. Raw predictors and deterministic
  packaging are separated explicitly. Lower is better.}
  \label{tab:official_results}
  \footnotesize
  \setlength{\tabcolsep}{1.4pt}
  \begin{tabular}{llrrr}
    \toprule
    System & Type & Final & Range & Angle \\
    \midrule
    Official baseline & ref. & 0.633957 & 0.988067 & 0.279847 \\
    Reloc3r 1 epoch & learned & 0.003188 & 0.002528 & 0.003849 \\
    Metric-aware RPR & learned & 0.003078 & 0.002412 & 0.003744 \\
    MDHR & learned & \textbf{0.002460} & \textbf{0.002274} & 0.002646 \\
    PAAER & learned & 0.002514 & 0.002392 & \textbf{0.002636} \\
    \midrule
    PACE final & package & 0.001874 & 0.001330 & 0.002419 \\
    \bottomrule
  \end{tabular}
\end{table}

The released reconstruction reads the three raw official prediction files,
applies Eqs.~\eqref{eq:rsf}--\eqref{eq:structured_projection}, writes the
required angle--range order, and verifies artifact hashes. This reconstructs
the released submission package; stochastic retraining is not claimed to
reproduce identical checkpoint bytes.

\section{Discussion and Limitations}
\label{sec:discussion}

The results support a bounded conclusion: PairUAV benefits from a shared
image-pair representation, but its polar axes need not inherit the same
decoder evidence or training state. Matched depth controls provide the direct
architectural evidence, while representation, trajectory, occlusion, and tail
analyses describe complementary aspects of axis-conditioned reliability. MDHR
and PAAER instantiate this principle without claiming hard factor
disentanglement or complete gradient isolation. Range fusion, lattice rounding,
and support projection improve the submitted package but remain deterministic,
PairUAV-specific operations rather than learned contributions. This boundary
also governs the official comparison: MDHR has lower final and range relative
errors, whereas PAAER improves the angle term by only 0.000010. We therefore
treat them as complementary asymmetric readouts rather than a sequence in which
PAAER supersedes MDHR. Likewise, the tail-weighted continuation tests whether a
diagnosed high-error range mode is repairable; it is not a same-budget
architecture ablation.

Several limitations constrain generalization. Some adaptation results
come from different bounded protocols; the patch-occlusion panel contains four
selected cases; and checkpoint winners are label-assisted diagnostics rather
than an inference policy. The observed axis-dependent tails can reflect visual
evidence, optimization, and the evaluator's relative-error denominators.
Moreover, the $2^\circ$ heading lattice and public range support may not exist
in another benchmark. The range-fusion coefficients were selected during
challenge development from aggregate leaderboard feedback, so their gain should
be read as challenge engineering rather than a transferable learned rule. The
released artifacts guarantee deterministic reconstruction from fixed raw
predictions and verified hashes, not stochastic retraining to identical
checkpoint bytes. Future work should replace post-hoc fusion and snapping with
trainable per-axis reliability. A controlled follow-up should compare identical
depth probes and gradient-isolated paths under matched initialization and budget,
and report circular heading MAE and absolute range MAE alongside the official
relative errors, stratified by denominator magnitude. Structured output could
then be learned through classification-plus-residual heads rather than imposed
only after inference. The central design question remains broader than the
package: in a shared pairwise representation, each predicted coordinate should
read the evidence on which it is reliable.

\bibliographystyle{unsrtnat}
\bibliography{refs}

\end{document}